\documentclass[10pt,twocolumn]{article} 
\usepackage{simpleConference}
\usepackage{times}
\usepackage{graphicx}
\usepackage{amssymb}
\usepackage{url,hyperref}

\usepackage{times}
\usepackage{epsfig}
\usepackage{graphicx}
\usepackage{amsmath}
\usepackage{amssymb}

\usepackage{algorithmic}
\usepackage[dvipsnames]{xcolor}
\usepackage{algorithm}
\usepackage{subfigure}
\usepackage{amsmath}
\usepackage{booktabs}
\usepackage{ragged2e}

\usepackage{bbm}
\usepackage{ragged2e}
\usepackage{multirow}
\usepackage{mathtools}
\usepackage{gensymb}
\usepackage{tabularx}
\usepackage{makecell}
\usepackage{subfiles}

\usepackage[font=md]{caption}

\begin{document}

\title{The Devils in the Point Clouds: Studying the Robustness of Point Cloud Convolutions}

\author{Xingyi Li, Wenxuan Wu, Xiaoli Z. Fern, and Li Fuxin\\
Oregon State University\\
{\tt\small \{lixin, wuwen, xfern, lif\}@oregonstate.edu}
}

\maketitle

\begin{abstract}
Recently, there has been a significant interest in performing convolution over irregularly sampled point clouds. Since point clouds are very different  from regular raster images, it is imperative to study the generalization of the convolution networks more closely, especially their robustness under variations in scale and rotations of the input data. This paper investigates different variants of PointConv, a convolution network on point clouds, to examine their robustness to input scale and rotation changes. Of the variants we explored, two are novel and generated significant improvements. The first is replacing the multilayer perceptron based weight function with much simpler third degree polynomials, together with a Sobolev norm regularization. Secondly, for 3D datasets, we derive a novel viewpoint-invariant descriptor by utilizing 3D geometric properties as the input to PointConv, in addition to the regular 3D coordinates. We have also explored choices of activation functions, neighborhood, and subsampling methods. Experiments are conducted on the 2D MNIST \& CIFAR-10 datasets as well as the 3D SemanticKITTI \& ScanNet datasets. Results reveal that on 2D, using third degree polynomials greatly improves PointConv's robustness to scale changes and rotations, even surpassing traditional 2D CNNs for the MNIST dataset. On 3D datasets, the novel viewpoint-invariant descriptor significantly improves the performance as well as robustness of PointConv. We achieve the state-of-the-art semantic segmentation performance on the SemanticKITTI dataset, as well as comparable performance with the current highest framework on the ScanNet dataset among point-based approaches.
\end{abstract}

\section{Introduction}
\label{sec:intro}
Convolution is one of the most fundamental concepts in deep learning. Convolutional neural networks (CNNs) have redefined the state-of-the-art for almost every task in computer vision. In order to transfer such successes from 2D images to the 3D world, there is a significant body of work aiming to develop the convolution operation on 3D point clouds. This is essential to many applications such as autonomous driving and virtual/augmented reality.

PointConv~\cite{Wu_2019_CVPR} seems to be one of the promising efforts. Similar to earlier work~\cite{8099494,NIPS2016_6578,dgcnn,hermosilla2018mccnn,Wang_2018_CVPR,Xu_2018_ECCV}, PointConv utilizes a multi-layer perceptron (MLP) to learn the convolution weights on each point implicitly as a nonlinear transformation from the point coordinates, basically a Monte Carlo discretization of the continuous convolution operator on irregular point clouds, but with the efficient version of PointConv it now can scale to modern deep networks with dozens of layers. PointConv is permutation-invariant and translation-invariant, and to our best knowledge is the only approach that achieved equivalent performance to 2D CNNs on images as well as having one of the highest performances on 3D benchmarks.

\begin{figure*}[t!]
    \vskip -0.15in
    \centering
    \subfigure[]
    {\includegraphics[width=0.3\textwidth]{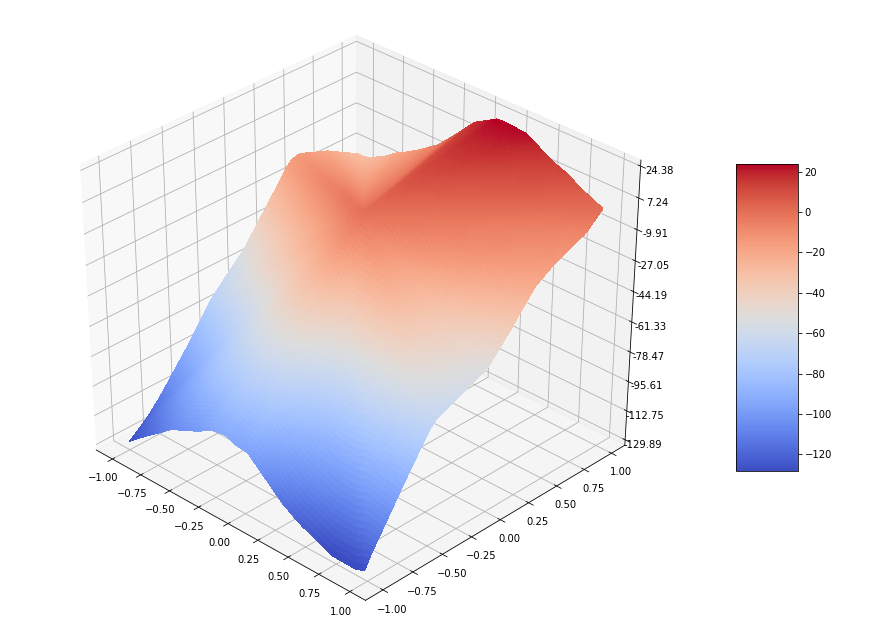}} \label{fig:MNISTFilter}
    \subfigure[]
    {\includegraphics[width=0.3\textwidth]{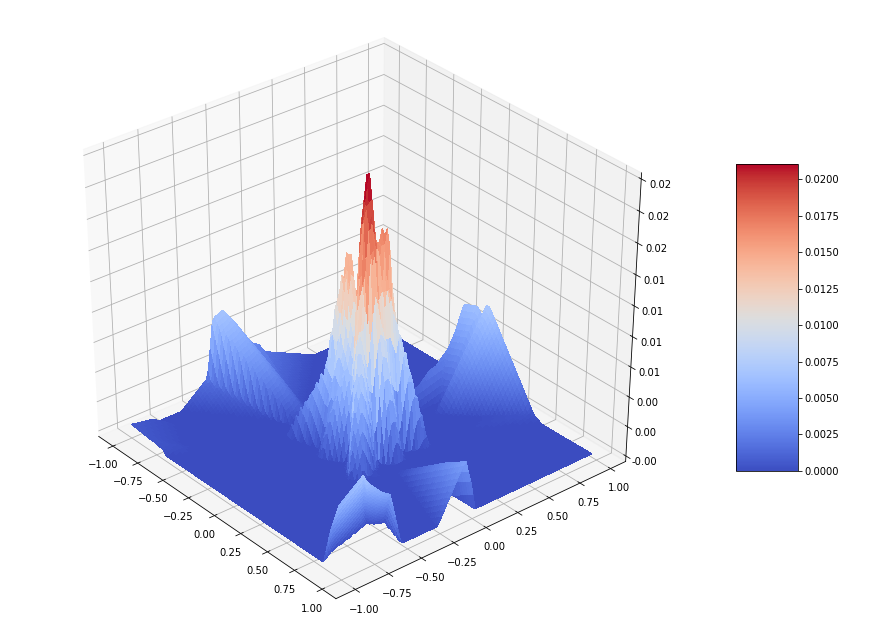}} \label{fig:FasterRcnnFilter}
    \subfigure[]
    {\includegraphics[width=0.3\textwidth]{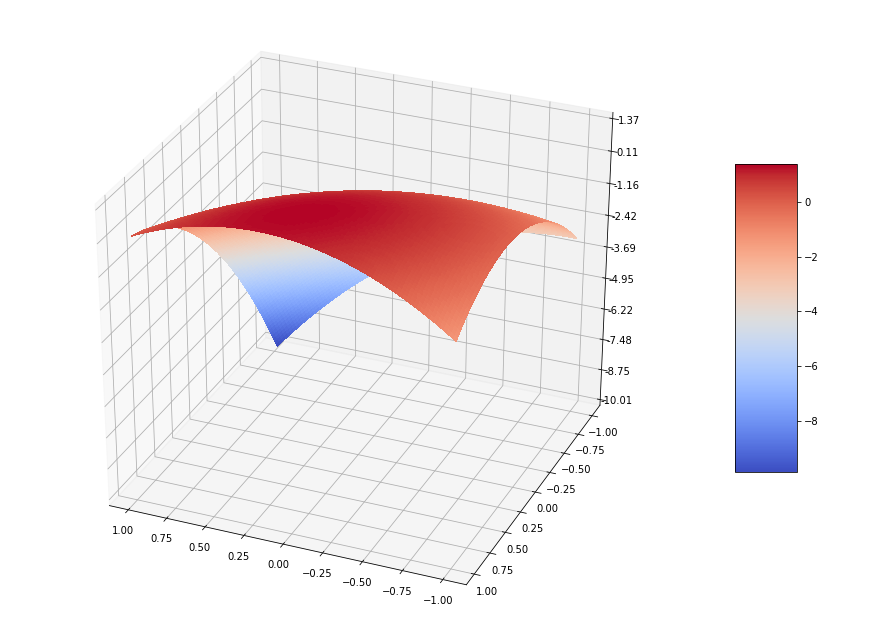}} \label{fig:ploy3Fig}
    \caption{Examples of learned weight functions from  PointConv: (a) MLP-based PointConv trained on MNIST; (b) MLP-based PointConv trained with Faster R-CNN; (c) Sobolev-regularized cubic polynomial (Best Viewed in Color)}
    \label{fig:filterPlot}
    \vskip -0.05in
\end{figure*}

Networks based on point clouds introduce a new complication on the neighborhoods used in convolution. In 2D images, we are accustomed to having fixed-size neighborhoods such as $3\times 3$ or $5 \times 5$. PointConv and other point-based  networks instead adopt k-nearest neighbors (kNN), which may potentially make it harder for point cloud networks to  generalize from training locations to testing locations, as sufficiently smooth weight functions need to be learned. Usually, point cloud networks augment the data by randomly jittering point locations, but such jittering only provides local generalization. We attempted to plot one of the typical learnt weight function on MNIST and on a faster-RCNN detector as shown in Fig.~\ref{fig:filterPlot} (a-b). As one can see, due to the nonlinearity in the weight functions, PointConv could potentially generalize poorly if the testing neighborhoods are substantially different from those of the training. Indeed, even in 2D images we rely on re-scaling of all the images to the same scale to avoid this generalization issue. Such a simple shortcut is, however, unlikely to suffice for point clouds as each kNN neighborhood may be significantly different from others in terms of scale.

In this paper we study empirically the generalization of PointConv under scale changes (resulting in different sampling densities) and rotations, which would induce very different neighborhoods between training and testing. The basic methodology is to train the network with a certain set of scales and rotations, and test it on out-of-sample scales/rotations that are significantly different from the training ones. Experiments are done both on the 2D MNIST \cite{mnist} \& CIFAR-10 \cite{Cifar} datasets, and the 3D SemanticKITTI \cite{behley2019iccv} \& ScanNet v2 \cite{dai2017scannet} datasets. Multiple design choices are tested, including different neighborhood selection methods, activation functions, input feature transformations and regularization methods to examine their impact on generalization under scale changes and rotations.

From the experiments, we identify two strategies that have not been applied to point cloud networks to be the most effective. In 2D images, we propose to utilize cubic polynomials as the weight functions, with a Sobolev norm regularization similar to thin-plate splines. This restricts the flexibility of the weight functions (Fig.~\ref{fig:filterPlot}) and improves generalization. We additionally find that using an $\epsilon$-ball neighborhood is more robust than kNN in 2D. With these improvements, we have found PointConv to be more robust than traditional raster CNNs on scale changes and rotations, which suggests a potential of applying PointConv on 2D images for the sake of robustness in future work. 

For 3D point clouds, we introduce a novel viewpoint-invariant (VI) feature transformation to the 3D coordinates. The results show that our novel feature transformation is not only rotation-invariant, but also robust to neighborhood size changes and achieves significantly better generalization results than simply using 3D coordinates as input. VI enables coordinates to generalize to neighborhood of different sizes, hence we have found that $\epsilon$-ball neighborhoods become no longer necessary on top of VI. This is welcome news since $\epsilon$-ball neighborhoods (e.g. utilized commonly in KPConv~\cite{thomas2019KPConv}) are more expensive to compute than kNN neighborhoods.

\section{Related Work}
\noindent \textbf{Volumetric and Projection-based Approaches} 
A direct extension from convolution in 2D raster images to 3D is to compute convolution on volumetric grids~\cite{wu20153d,maturana2015voxnet,OctNet,Wang-2017-OCNN}. In densely sampled point clouds, sparse volumetric convolutions such as MinkowskiNet \cite{Minkowski} and Submanifold sparse convolution \cite{Submanifold} currently obtain the best performance. However, they depend heavily on being able to locate enough points in a local volumetric neighborhood of each point, hence are difficult to apply to cases where the sampling density of point clouds is low or especially uneven (e.g. LIDAR).
Some other approaches that project point clouds onto multi-view 2D images \cite{su15mvcnn, qi2016volumetric, Lin_2020_CVPR} or lattice space \cite{su18splatnet} may suffer from the same issue.
 
\noindent \textbf{Point-based Approaches} \cite{qi2016pointnet} first attempted to directly work on point clouds, and PointNet++ \cite{qi2017pointnetplusplus} improved it by adding a hierarchical structure. Following PointNet++, some other studies also attempted to utilize hierarchical architectures to aggregate information from neighbor points with MLPs \cite{conf/cvpr/LiCL18, liu2019point2sequence}. PointCNN \cite{NIPS2018_7362} utilized a learned $X$-transformation to weight the local features and reorder them simultaneously. FeaStNet \cite{verma:hal-01540389} utilize a soft-assignment matrix to generalize traditional convolution on point clouds. Flex-convolution \cite{accv2018/Groh} introduced a convolution layer for arbitrary metric spaces, along with an efficient GPU implementation. A-CNN \cite{komarichev2019acnn} proposed a annular convolution that assigned kernel weights for neighbour points based on their distances to the center point. PointWeb \cite{zhao2019pointweb} considered every pair of points within the neighborhood for extracting contextual features. A-SCN \cite{Xie_2018_CVPR} adopt the dot-product self-attention mechanism from \cite{NIPS2017_7181} to propagate features. PointASNL \cite{Yan_2020_CVPR} proposed an adaptive sampling strategy to avoid outliers before extracting local and global features from each point cloud.

Generally, convolutional approaches on point clouds performed better than the approaches listed above.  \cite{8099494,NIPS2016_6578,dgcnn,hermosilla2018mccnn,Wang_2018_CVPR,Xu_2018_ECCV} proposed to learn discretizations of continuous convolutional filters. \cite{NIPS2016_6578} utilized a side network to generate weights for 2D convolutional kernels. \cite{8099494} generalized it to 3D point clouds, and \cite{Wang_2018_CVPR} further extends it to segmentation tasks \cite{Wang_2018_CVPR}, along with an efficient version. However, the efficient version in~\cite{Wang_2018_CVPR} only achieves depthwise convolution rather than full convolution. EdgeConv~\cite{dgcnn} encodes pairwise features between a neighbor point and the center point through MLPs. \cite{hermosilla2018mccnn} takes densities into account. Pointwise CNN \cite{hua-pointwise-cvpr18} located kernel weights for predefined voxel bins, so it was not flexible. SpiderCNN \cite{Xu_2018_ECCV} proposed a polynomial weight function, which we experiment with in this paper. However, they did not utilize regularization to control the smoothness. The formulation of PointConv~\cite{Wu_2019_CVPR} is mathematically similar to \cite{hermosilla2018mccnn}, and it encompasses \cite{8099494,dgcnn} and  \cite{Wang_2018_CVPR}, since those can be viewed as special cases of PointConv, removing some of the components (e.g. density, full convolution). The main contribution in PointConv \cite{Wu_2019_CVPR} is an efficient variant that does not explicitly generate weight functions, but implicitly so by directly computing the convolution results between the weights and the input features. Such a variant removed the memory requirement to store the weights and networks, allowing for scaling up to the ``modern" deep network size, e.g. dozens of layers with hundreds of filters per layer. It is also the only paper showing results on CIFAR-10 matching those of a 2D CNN of the same structure.

The main competitive point-based convolutional approach to PointConv is KPConv \cite{thomas2019KPConv}. In KPConv, convolution weights are generated as kernel functions between each point and anchor points, points in the 3D space that are pre-specified as parameters for each layer separately. KPConv enjoys nice performance due to the smooth and well-regularized kernel formulation, but it introduces significantly more parameters in the specification of anchor points and their $\epsilon$-ball neighborhoods are computationally costly. Similar to KPConv, PCNN \cite{atzmon2018point} also assign anchor points with kernel weights, but it does not take neighbor points into account for convolution.

\noindent \textbf{Scale and rotation invariance in convolution}
We did not find significant amount of related work in studying scale and rotation robustness on point clouds. \cite{spn3dpointclouds,yuan2018iterative} build a spatial transformer side network (STN) to learn global transformations on input point clouds. The main difference comparing with our work is that we aim to improve the robustness against transformations through directly working on the network itself, instead of designing additional structures to accommodate them.

A significant amount of work have been published in 2D CNNs on scale and rotation invariance. A standard approach has been data augmentation \cite{Simonyan2014VeryDC,He2014,ArtAugmentation, TI-POOLING, Cheng_2016_CVPR, henriques17warped}, where the training set is augmented by including objects with random rescaling or rotations. A group of studies attempted to integrate deep CNNs with side-networks \cite{Lin2017,Zhou_2018_CVPR,Location-Aware,DBLP:journals/corr/abs-1712-02408,Zhang2017ScaleAdaptiveCF,DBLP:journals/corr/abs-1808-04974} or attention modules \cite{wang2017residual,sharma2015action}. \cite{Zhou2017ORN} convolved the input with several rotated versions of the same CNN filter before feeding to the pooling layer. Some techniques proposed to learn transformations directly \cite{jaderberg2015spatial,lin2017inverse} on the input or intermediate outputs from convolutional layers in a deep network ~\cite{Dai2017,DBLP:journals/corr/abs-1710-09829}. There has also been interest in combining group concepts with CNNs to encode scale and rotate transformations \cite{cohen2016group,byravan2017se3,cohen2017steerable}.

\noindent \textbf{Non-deep Approaches}
\cite{Rahul} encodes relations between local 3D surface patches as well as global patches in a viewpoint invariant manner, which is a good hint for this work.

\section{Methodology}
The main goal of this work is to investigate how to best enable the learned weight function to generalize from known local locations to unseen ones, to make PointConv~\cite{Wu_2019_CVPR} based networks more robust against unseen scales and rotations of objects. Toward this, we attempted three methods. First, we replace kNN with $\epsilon$-ball based neighbor search to unify the receptive field for each PointConv layer. Next, we introduce a much simpler hypothesis space of third degree polynomials to replace MLP for weight functions to avoid overfitting. To further enforce the smoothness of this hypothesis space, we utilize the Sobolev norm for thin-plate splines as a regularizer. Finally, for 3D point clouds, we introduce a viewpoint-invariant (VI) descriptor for the MLP that utilizes geometric properties of the data to be less sensitive to local scale and rotation changes.

Below we will first introduce   PointConv \cite{Wu_2019_CVPR}, followed by the description of $\epsilon$-ball based neighbor search, third degree polynomials, and the VI descriptor for the weight function, respectively.

\begin{figure*}[h!]
    \centering
    \subfigure[]
    {\includegraphics[width=0.52\textwidth]{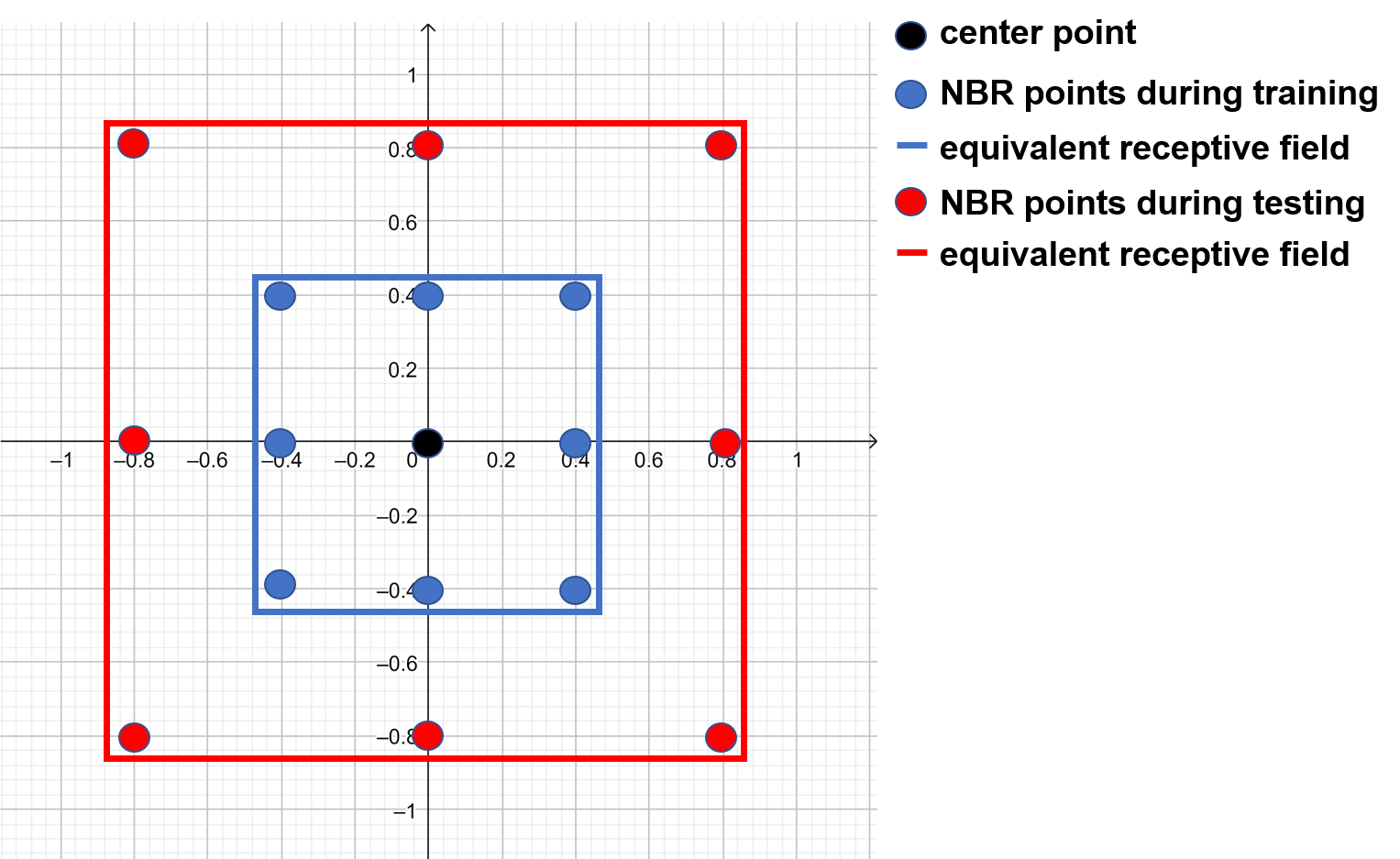}} \label{fig:eballNBR}
    \subfigure[]
    {\includegraphics[width=0.4\textwidth]{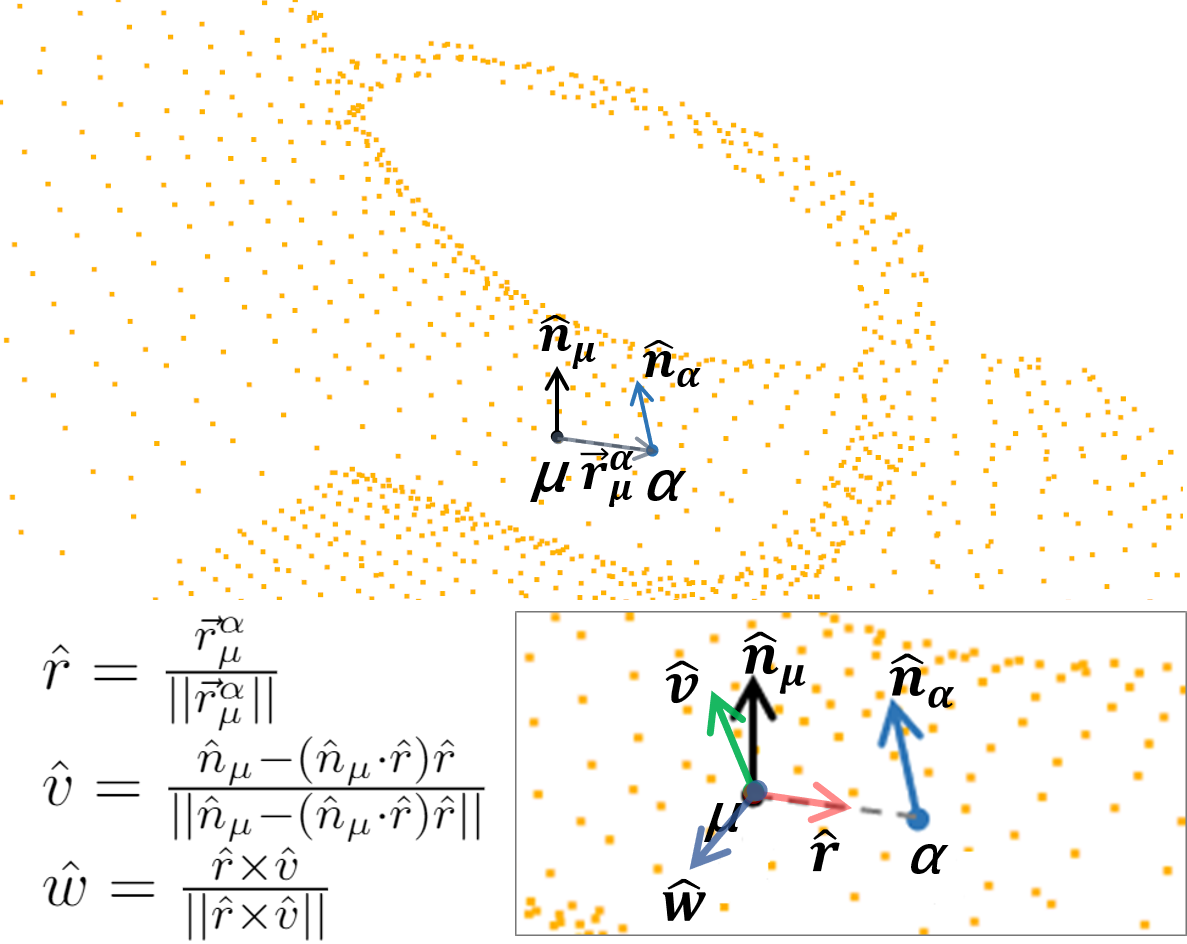}} \label{fig:viewInvariant}
    \caption{(a) We perform robustness experiments of PointConv on 2D images where the training kNN neighborhood is significantly different from the testing; (b) For a given local center point $p_\mu$ and $p_\alpha \in N_\epsilon(p_\mu)$ for a pair of points, a set of viewpoint-agnostic basis $(\vec{r}, \vec{w},\vec{v})$ can be generated from $\vec{r}_\mu^\alpha$ and $n_\mu$ with the Gram-Schmidt process, and viewpoint-invariant features such as the angles between $\vec{n_\mu}$ and $\vec{v}$ can be extracted from them}
    \label{fig:approach}
    \vskip -0.05in
\end{figure*}

\subsection{Background: PointConv}
A point cloud can be denoted as a set of $3D$ points $P=\{p_1,p_2,...,p_n\}$, where each point $p_i$ contains a position vector $(x,y,z)\in R^3$ as well as a feature vector (RGB color, surface normal, etc.). A line of work including PointConv generalizes the convolution operation to point clouds based on discretizations of continuous 3D convolutions~\cite{8099494,hermosilla2018mccnn,dgcnn,Wu_2019_CVPR}. For a center point $p_{xyz}=(x,y,z)$, its PointConv is defined by:

\begin{equation}
\small
\label{eq:PConvEq}
\begin{multlined}
    PointConv(S,W,F)_{xyz}=\\
    \sum_{(\delta_x,\delta_y,\delta_z)\in G}S(\delta_x,\delta_y,\delta_z)W(\delta_x,\delta_y,\delta_z)F(x+\delta_x,y+\delta_y,z+\delta_z)
\end{multlined}
\end{equation}
where $(\delta_x,\delta_y,\delta_z)$ denote the coordinate offsets for a point in $p_{xyz}$'s local neighborhood $G$, usually located by kNN.  $F(x+\delta_x,y+\delta_y,z+\delta_z)$ represents the feature of the point, and $W(\delta_x,\delta_y,\delta_z)$ is the convolution kernel generating the weights for convolution and is approximated by an MLP, called \textit{WeightNet} in \cite{Wu_2019_CVPR}. Finally, $S(\delta_x,\delta_y,\delta_z)$ represents the inverse local density to balance the impact of non-uniform sampling of the point clouds.

PointConv uses an efficient approach to avoid the computation of the function values of $W$ at each point, which is extremely memory-intensive. The computation approach does not change the final output of eq.\ (\ref{eq:PConvEq}), hence we omit it. A deep network can be built from PointConv layers similar to 2D convolutions. For stride-2 convolution/pooling, one can just subsample the point clouds~\cite{qi2017pointnetplusplus}. \cite{Wu_2019_CVPR} also provides a deconvolution/upsampling approach to increase the resolution of point clouds. Hence, classification and semantic segmentation tasks can be solved with PointConv networks. It is also straightforward to incorporate other commonly used 2D convolution operations, e.g.\ residual connections. Dilated convolution can be approximated by first sampling a larger kNN neighborhood, and then subsampling from the neighborhood.

\subsection{$\epsilon$-ball neighbor search and activations}
\label{section:eballNeighor search}
The neighborhood $G$ in PointConv is usually defined by kNN. Fig.~\ref{fig:approach} (a) illustrates the robustness issue for K-nearest neighbors search. Namely, the equivalent receptive field for a sparse point cloud is much larger than the one for a densely distributed point cloud. If trained only on dense (high resolution) point clouds, the learned weight function may not generalize well to much larger (unseen) receptive fields when dealing with sparse point clouds during testing.

An $\epsilon$-ball based neighborhood (e.g. commonly used in KPConv ~\cite{thomas2019KPConv}) on the other hand would be robust to different sampling rates. For a point $p_i$, denote  $N_\epsilon(p_i) = \{p_j\in P|d(p_i,p_j)<\epsilon\}$ as its $\epsilon$-ball neighborhood. To ease the computation burden, we (randomly) select at most $K$ neighbors from $N_\epsilon(p_i)$. The actual chosen neighbors from $N_\epsilon(p_i)$ are denoted as $C_\epsilon(p_i,K)$. Compared with kNN, $\epsilon$-ball neighborhood retains the maximal distance of the neighbors w.r.t.  the center point. Since different $\epsilon$-balls may contain different number of neighbors, we replace the normalizer $S(\delta_x,\delta_y,\delta_z)$ in equation \ref{eq:PConvEq} with $\frac{1}{|C_\epsilon(p_i,K)|}$. Note that the flexibility of the PointConv framework allows for variable number of neighbors in each neighborhood.

We also investigate the robustness over several different activations in the MLP-based WeightNet, such as ReLU \cite{ReLU}, SeLU \cite{Selu}, Leaky ReLU \cite{leaky}, and Sine $(\sin(x))$. Sine is included because its connections to random Fourier features~\cite{rahimi2008random}. In  ~\cite{rahimi2008random}, it was proved that a basis of $\cos(\mathbf{Wx}+\mathbf{b})$ with random $\mathbf{W}$ and $\mathbf{b}$ could be a universal function approximator. Hence we thought learned $\mathbf{W}$ and $\mathbf{b}$s could improve over the pure random one. Empirically we have found that using sine worked better than cosine, maybe due to the fact that $sin(0) = 0$ hence it does not introduce additional constants.

\subsection{Convolutional Kernels as Cubic Polynomials}
The set of functions MLPs can represent is very large, which may introduce overfitting to the training point locations. Hence, we experiment with much simpler weight functions $W(\delta_x, \delta_y, \delta_z)$ 
in the form of cubic polynomials of $(x,y,z)$. This was investigated in ~\cite{Xu_2018_ECCV}, however with some arbitrary additional functions multiplied that actually reduced the performance on 2D~\cite{Wu_2019_CVPR}. We utilize a plain version, with proper weights to regularize for smoothness. In 2D, this results in a feature space:
\begin{equation}
\small
\label{eq:2DPoly}
    \phi(x,y) = [x^3,y^3,\sqrt{3}x^2y,\sqrt{3}xy^2,\sqrt{3}x^2,\sqrt{3}y^2,\sqrt{6}xy,x,y,1]
\end{equation}
and in 3D:
\begin{equation*}
\label{eq:3DPoly}
\begin{split}
    \phi(x,y,z) = [x^3,y^3,z^3,\sqrt{3}x^2y,\sqrt{3}x^2z,\sqrt{3}xy^2,\sqrt{3}y^2z, \\\sqrt{3}xz^2,
    \sqrt{3}yz^2,\sqrt{3}x^2,\sqrt{3}y^2,\sqrt{3}z^2, \sqrt{6}xyz,\sqrt{6}xy,\\\sqrt{6}xz,\sqrt{6}yz,\sqrt{3}x, 
    \sqrt{3}y,\sqrt{3}z,1]
\end{split}
\end{equation*}

The coefficients of each term ensures that an $L_2$ regularization on the feature descriptors (excluding linear and constant terms) correspond to regularizing with the Sobolev $S_2$ norm:
\begin{equation}
\label{eq:2DPolySmooth}
    ||f||_{s_2}^2 = \lambda \int\int[(\frac{\partial^2f}{\partial x^2})^2+2(\frac{\partial^2f}{\partial x\partial y})^2+(\frac{\partial^2f}{\partial y^2})^2]dxdy
\end{equation}
in 2D, where $\lambda$ is a parameter. The 3D form can be written similarly. Sobolev-norm regularizations are commonly used in thin-plate splines~\cite{Wahba90a} but to our knowledge they have not been used in point cloud networks in the past.
Note that the choice of a third degree polynomial is common in the smoothing splines community. Quadratic functions have undesirable symmetries and fourth-order polynomials introduce too many terms, e.g. $35$ terms for a 3D space.

\subsection{A Viewpoint-Invariant Input Descriptor}
PointConv relies on the $(x,y,z)$ coordinates to compute the weights, which is sensitive to the rotation of the object as well as the sampling rate of the point clouds. We hypothesize that by using viewpoint-invariant descriptors for the weight function, we can achieve better generalization.

We develop a viewpoint-invariant (VI) descriptor for each point $p_\alpha$ in a local neighborhood as an $8$-dimensional vector utilizing its geometric relationship with the center point $p_\mu$. Denote the surface normal of $p_\mu$ as $\hat{n}_\mu$ and its difference with  $p_\alpha$ as $\vec{r}_\mu^\alpha = p_\alpha - p_\mu$. When the scene is rotated, the angle between $\hat{n}_\mu$ and $\vec{r}_\mu^\alpha$ 
remains the same. To discriminate between different directions, we generate an orthonormal basis from $\{\hat{n}_\mu, \vec{r}_\mu^\alpha\}$ and compute the angles between $\hat{n}_\mu$, $\vec{r}_\mu^\alpha$ and $\hat{n}_\alpha$ as well as the projection lengths of $\hat{n}_\alpha$ and $\hat{n}_\mu$ onto the orthonormal basis. With a global rotation of the scene, 
the basis and normal vectors are identically rotated. Hence, our descriptor is rotation invariant and provides a complete characterization of the vectors $\hat{n}_\mu$, $\vec{r}_\mu^\alpha$ and $\hat{n}_\alpha$.

Formally, we utilize the Gram-Schmidt process on $\{\vec{r}_\mu^\alpha, \hat{n}_\mu\}$ to generate a 3D basis $\{\hat{r},\hat{v}, \hat{w}\}$ where $\hat{r} = \frac{\vec{r}_\mu^\alpha}{||\vec{r}_\mu^\alpha||}$, $\hat{v} = \frac{\hat{n}_\mu - (\hat{r}^\top \hat{n}_\mu)  \hat{r}}{ \sqrt{1 - (\hat{r}^\top \hat{n}_\mu)^2 }}$ is orthonormal to $\hat{r}$, and $\hat{w}$ is derived by $\hat{r} \times \hat{v}$, the outer product of $\hat{r}$ and $\hat{v}$, as illustrated in Fig.~\ref{fig:approach}. Note that this basis is seldom degenerate, because it is unlikely for $\hat{n}_\mu$ and $\vec{r}_\mu^\alpha$ to be collinear in 3D surface point clouds.

With the basis defined, for each point $p_\alpha$ in the neighborhood of $p_\mu$, we extract the following viewpoint invariant feature vector:
\begin{equation}
\label{eq:viewPointInvar}
    \begin{split}
    \beta_\mu^\alpha = [\hat{n}_\alpha \cdot  \hat{n}_\mu,\frac{\vec{r}_\mu^\alpha \cdot \hat{n}_\mu}{\|\vec{r}_\mu^\alpha\|},\frac{\vec{r}_\mu^\alpha \cdot \hat{n}_\alpha}{\|\vec{r}_\mu^\alpha\|}, \hat{n}_\alpha \cdot \hat{v}, \hat{n}_\alpha \cdot \hat{w},\\
    \vec{r}_\mu^\alpha \cdot \hat{n}_\mu, \vec{r}_\mu^\alpha \cdot (\hat{n}_\alpha \times \hat{n}_\mu), ||\vec{r}_\mu^\alpha||]
    \end{split}
\end{equation}

where  $\times$ represents the cross product. 
The first $5$ features are both scale and rotation invariant, and the last $3$ features are rotation invariant only. We believe that a weight function with this input will be invariant to rotation and more robust against scale changes.
We also anticipate this to alleviate the need for rotation-based data augmentation.

\section{Experiments}
\subsection{MNIST}
\label{PConv_MNIST_Experiments}
MNIST contains $60,000$ handwritten digits in the training
set and $10,000$ digits in the test set from 10 distinct classes. The original resolution of each image is $28 \times 28$. We trained a 4-layer network. Table \ref{table:MNIST arch} illustrate the general architecture for our proposed framework and all other baselines. They only differ in the structures for convolutional layers. 
The $\epsilon$ is fixed to $\frac{1}{10}$ for Conv1 in Table \ref{table:MNIST arch}, $\frac{1}{5}$ for Conv2, and $\frac{1}{2}$ for Conv3 as well as Conv4. The numbers of output points from FPS for Subsampling1 Subsampling2 layer in Table \ref{table:MNIST arch} are $196$ and $49$ respectively. The coefficient of the Sobolev norm regularization is set to $10^{-6}$ or $10^{-5}$ for our proposed approach. For other ablation variants, it was tuned to the optimal one. 
We utilize cross entropy as the loss function. The batch size is $60$ and the optimizer is Adam with learning rate $0.001$. For traditional 2D CNN baselines, we adopt max pooling with $2\times 2$ stride for the sub-sampling layer. Global average pooling is used in the last layer to account for different input resolutions. For point cloud sub-sampling, we attempt two different methods. The first one is named as PC-2D subsampling, which achieves the equivalent number of sampled points as 2D max-pooling with stride $2\times 2$, but in point cloud formats. The next one is Farthest Point Sampling (FPS), a commonly used sampling approach in point cloud networks~\cite{qi2017pointnetplusplus,Wu_2019_CVPR}.

\begin{table*}[h!]
\begin{center}
\small
\begin{tabular}{ c | c } 
 \hline
 \textbf{Layer name} & \textbf{Layer description}  \\ 
 \hline
 Conv1 & $3\times3$ conv. (or PointConv) 64 w/ ReLU\\
 \hline
 Subsampling1 & max-pooling with stride $2\times 2$ (or farthest point sampling)\\
 \hline
 Conv2 & $3\times3$ conv. (or PointConv) 128 w/ ReLU\\
 \hline
 Subsampling2 & max-pooling with stride $2\times 2$ (or farthest point sampling)\\
 \hline
 Conv3 & $3\times3$ conv. (or PointConv) 128 w/ ReLU\\
 \hline
 Conv4 & $3\times3$ conv. (or PointConv) 128 w/ ReLU\\
 \hline
Pooling & global average pooling\\
 \hline
FC-10 & fully connected layer\\
 \hline
\multicolumn{2}{c}{Softmax layer}\\
 \hline
\end{tabular}
\end{center}
\caption{The general network architecture for MNIST experiments. Every convolutional layer is followed by a ReLU layer.}
\label{table:MNIST arch}
\end{table*}

To convert a 2D raster image with the resolution of $m\times n$ to a point cloud, we generate a point for each pixel with normalized coordinates $p_i(x,y) = (\frac{m_i}{m},\frac{n_i}{n}) \in R^2$ as its spatial location. Normalization constrains the input range of each coordinate for the weight function to $[0,1]$. Thus, images of different scales are converted to point clouds with different sampling densities.
The original RGB feature is also normalized to $[0,1]$. 

Three 2D CNN baselines are considered. The first one is exactly the same structure as the PointConv network. The other two utilize the Deformable CNN~\cite{Dai2017}, and CapsNet~\cite{DBLP:journals/corr/abs-1710-09829} respectively. Both claimed to be robust to scale/rotation changes.
Every baseline is tuned to the optimal performance based on the validation accuracy.

\noindent\textbf{Robustness to Scaling}
The first experiment we conduct is to train with images of limited known scales and test on larger/smaller objects outside of the known scales. To construct the training set, all MNIST images are rescaled to $20\times20$, $28\times28$, and $36\times36$, respectively. The validation set is built by rescaling the original test images to $24\times24$ and $32\times32$. We trained all baseline models on this new MNIST dataset, and tuned parameters on the validation set. Test accuracies are measured on images with $34\times34$, $38\times38$, $44\times44$, $56\times56$, $72\times72$, $18\times18$, $14\times14$, and $10\times10$ resolutions, which creates a discrepancy with training.

Results are shown in Table~\ref{table:MNISTFarthestSubResults} and \ref{table:MNISTGridSubResults}. PC-2D sampling is generally inferior to the furthest point subsampling. We note several takeaways:
\begin{itemize}
    \item Naive kNN-based PointConv is less robust to scale changes than conventional CNN. On most out-of-sample scales it does not generalize as well as the 2D CNNs.
    \item Cubic polynomials with Sobolev regularization outperforms MLP with any activation at most scales. Sobolev regularization improves robustness especially at small scales. 
    \item $\epsilon$-ball neighborhood is generally more robust than kNN neighborhood, especially at scales significantly different from the training.
    \item With appropriate design e.g. cubic polynomial WeightNet and $\epsilon$-ball neighborhood, PointConv is significantly more robust to scale changes than traditional 2D CNN.
\end{itemize}

This builds a case to use PointConv in 2D spaces with an $\epsilon$-ball neighborhood, and a cubic polynomial WeightNet with Sobolev regularizations. Especially, the significantly improved robustness w.r.t. 2D CNNs may justify the use of PointConv even for 2D raster images, if robustness is of concern.

\begin{table*}[t!]
\begin{center}
\small
\begin{tabular}{|@{\hskip 1pt}c@{\hskip 1pt}|@{\hskip 1pt}c@{\hskip 1pt}|@{\hskip 2pt}c@{\hskip 2pt}|@{\hskip 2pt} c@{\hskip 2pt} |@{\hskip 2pt} c@{\hskip 2pt} |@{\hskip 2pt} c@{\hskip 2pt} |@{\hskip 2pt} c@{\hskip 2pt} |@{\hskip 2pt} c@{\hskip 2pt} ||@{\hskip 2pt} c @{\hskip 2pt}| @{\hskip 2pt}c @{\hskip 2pt}| @{\hskip 2pt}c @{\hskip 2pt}| } 
 \hline
 \textbf{\# NBR} & \textbf{Neighborhood} & \textbf{WeightNet architecture}  & $34\times34$ & $38\times38$ & $44\times44$ & $56\times56$ & $72\times72$ & $18\times18$ & $14\times14$ & $10\times10$  \\ 
\hline
$16$ & $\epsilon$-ball & cubic P + Sobolev Reg.& $95.37\%$ & $94.35\%$ & $\mathbf{94.56}\%$ & $\mathbf{94.01}\%$ & $\mathbf{93.23}\%$ & $\mathbf{95.82}\%$ & $\mathbf{96.15}\%$ & $47.4\%$  \\ 
\hline
$9$ & $\epsilon$-ball & cubic P + Sobolev Reg. & $91.96\%$ & $93.31\%$ & $91.00\%$ & $92.20\%$ & $91.04\%$ & $95.51\%$ & $95.27\%$ & $\mathbf{68.83}\%$  \\ 
\hline
\hline
$9$ & kNN & cubic P + Sobolev Reg. & $93.78\%$ & $93.84\%$ & $90.21\%$ & $88.45\%$ & $69.41\%$ & $92.79\%$ & $84.56\%$ & $54.67\%$  \\ 
\hline
$9$ & kNN & cubic P  & $90.31\%$ & $94.35\%$ & $89.77\%$ & $90.42\%$ & $68.46\%$ & $91.37\%$ & $81.73\%$ & $24.02\%$  \\ 
\hline
$9$ & kNN & MLP w/ Sine   & $91.35\%$ & $92.91\%$ & $84.69\%$ & $79.45\%$ & $66.29\%$ & $86.06\%$ & $74.61\%$ & $31.6\%$  \\ 
\hline
$9$ & kNN & MLP w/ SeLU & $91.77\%$ & $90.52\%$ & $87.5\%$ & $82.66\%$ & $69.5\%$ & $91.74\%$ & $82.9\%$ & $38.18\%$  \\ 
\hline
$9$ & kNN & MLP w/ Leaky ReLU  & $95.57\%$ & $88.89\%$ & $80.69\%$ & $49.8\%$ & $37.89\%$ & $94.61\%$ & $81.22\%$ & $26.85\%$  \\ 
\hline
$9$ & kNN & MLP w/ ReLU  & $71.38\%$ & $75.04\%$ & $75.41\%$ & $51.33\%$ & $18.85\%$ & $69.23\%$ & $59.54\%$ & $29.88\%$  \\ 
\hline
\hline
\multicolumn{3}{|@{\hskip 1pt}c@{\hskip 2pt}|}{CNN} & $96.30\%$ & $95.68\%$ & $89.19\%$ & $51.23\%$ & $25.77\%$ & $75.6\%$ & $24.25\%$ & $11.2\%$  \\ 
\hline
\multicolumn{3}{|@{\hskip 1pt}c@{\hskip 2pt}|}{Deformable CNN} & $91.31\%$ & $16.05\%$ & $17.99\%$ & $11.56\%$ & $11.35\%$ & $74.40\%$ & $28.18\%$ & $14.74\%$  \\ 
\hline
\multicolumn{3}{|c|}{CapsNet} & $\mathbf{98.53}\%$ & $\mathbf{97.52}\%$ & $93.28\%$ & $43.25\%$ & $16.50\%$ & $80.07\%$ & $21.41\%$ & $8.65\%$  \\ 
\hline
\end{tabular}
\end{center}
\caption{MNIST performance comparison across different scales under the setting of farthest subsampling. P is short for polynomials, and NBR is short for neighbor. The first row indicates the resolution of test images.}
\label{table:MNISTFarthestSubResults}
\end{table*}

\begin{table*}[t!]
\begin{center}
\small
\begin{tabular}{|@{\hskip 1pt}c@{\hskip 1pt}|@{\hskip 1pt}c@{\hskip 1pt}|@{\hskip 2pt}c@{\hskip 2pt}|@{\hskip 2pt} c@{\hskip 2pt} |@{\hskip 2pt} c@{\hskip 2pt} |@{\hskip 2pt} c@{\hskip 2pt} |@{\hskip 2pt} c@{\hskip 2pt} |@{\hskip 2pt} c@{\hskip 2pt} ||@{\hskip 2pt} c @{\hskip 2pt}| @{\hskip 2pt}c @{\hskip 2pt}| @{\hskip 2pt}c @{\hskip 2pt}| }
 \hline
 \textbf{\# NBR} & \textbf{Neighborhood} & \textbf{WeightNet architecture}  & $34\times34$ & $38\times38$ & $44\times44$ & $56\times56$ & $72\times72$ & $18\times18$ & $14\times14$ & $10\times10$  \\ 
\hline
$16$ & $\epsilon$-ball & cubic P + Sobolev Reg. & $95.83\%$ & $96.21\%$ & $\mathbf{97.84}\%$ & $\mathbf{97.92}\%$ & $\mathbf{98.23}\%$ & $90.8\%$ & $64.47\%$ & $21.78\%$  \\ 
\hline
$9$ & $\epsilon$-ball & cubic P + Sobolev Reg. & $94.76\%$ & $95.58\%$ & $96.56\%$ & $96.89\%$ & $97.27\%$ & $\mathbf{93.51}\%$ & $\mathbf{88.20}\%$ & $\mathbf{23.59}\%$  \\ 
\hline
\hline
$9$ & kNN & cubic P + Sobolev Reg. & $97.64\%$ & $96.83\%$ & $95.92\%$ & $54.06\%$ & $23.99\%$ & $53.84\%$ & $17.98\%$ & $10.11\%$  \\ 
\hline
$9$ & kNN & cubic P & $97.85\%$ & $97.31\%$ & $95.38\%$ & $42.71\%$ & $13.77\%$ & $14.77\%$ & $12.02\%$ & $7.28\%$  \\ 
\hline
$9$ & kNN & MLP w/ Sine  & $95.76\%$ & $94.35\%$ & $94.47\%$ & $47.26\%$ & $19.02\%$ & $76.5\%$ & $30.01\%$ & $10.86\%$  \\ 
\hline
$9$ & kNN & MLP w/ SeLU & $92.03\%$ & $82.49\%$ & $56.90\%$ & $17.09\%$ & $6.01\%$ & $33.78\%$ & $10.51\%$ & $11.59\%$  \\ 
\hline
$9$ & kNN & MLP w/ Leaky ReLU & $66.27\%$ & $24.39\%$ & $13.69\%$ & $10.11\%$ & $10.1\%$ & $65.66\%$ & $16.50\%$ & $8.11\%$  \\ 
\hline
$9$ & kNN & MLP w/ ReLU & $16.54\%$ & $16.66\%$ & $11.35\%$ & $11.35\%$ & $11.35\%$ & $10.63\%$ & $7.83\%$ & $6.43\%$  \\ 
\hline
\hline
\multicolumn{3}{|@{\hskip 1pt}c@{\hskip 2pt}|}{CNN} & $96.30\%$ & $95.68\%$ & $89.19\%$ & $51.23\%$ & $25.77\%$ & $75.6\%$ & $24.25\%$ & $11.2\%$  \\ 
\hline
\multicolumn{3}{|@{\hskip 1pt}c@{\hskip 2pt}|}{Deformable CNN} & $91.31\%$ & $16.05\%$ & $17.99\%$ & $11.56\%$ & $11.35\%$ & $74.40\%$ & $28.18\%$ & $14.74\%$  \\ 
\hline
\multicolumn{3}{|c|}{CapsNet}   & $\mathbf{98.53}\%$ & $\mathbf{97.52}\%$ & $93.28\%$ & $43.25\%$ & $16.50\%$ & $80.07\%$ & $21.41\%$ & $8.65\%$  \\ 
\hline
\end{tabular}
\end{center}
\caption{MNIST performance comparison across different scales under the setting of PC-2D subsampling. P is short for polynomials, and NBR is short for neighbor. The first row indicates the resolution of test images.}
\label{table:MNISTGridSubResults}
\end{table*}

\begin{table*}[t!]
\begin{center}
\small
\begin{tabular}{| c | c | c | c | c | c | c | c | c | } 
 \hline
 \textbf{\# NBR} & \textbf{Neighborhood} & \textbf{WeightNet architecture}  & $+10\degree$ & $-10\degree$ & $+20\degree$ & $-20\degree$ & $+40\degree$ & $-40\degree$ \\ 
\hline
$12$ & $\epsilon$-ball & cubic P + Sobolev Reg. & $96.93\%$ & $96.63\%$ & $96.2\%$ & $95.86\%$ & $85.27\%$ & $\mathbf{85.51}\%$ \\ 
\hline
$9$ & $\epsilon$-ball & cubic P + Sobolev Reg.  & $95.52\%$ & $95.58\%$ & $94.63\%$ & $94.55\%$ & $81.27\%$ & $84.27\%$ \\ 
\hline
\hline
$9$ & kNN & cubic P + Sobolev Reg. & $98.11\%$ & $97.68\%$ & $95.96\%$ & $97.2\%$ & $64.79\%$ & $63.71\%$ \\ 
\hline
$9$ & kNN & cubic P & $98.50\%$ & $98.37\%$ & $95.52\%$ & $97.45\%$ & $56.23\%$ & $61.75\%$ \\ 
\hline
$9$ & kNN & MLP w/ Sine  & $98.11\%$ & $97.65\%$ & $93.84\%$ & $93.95\%$ & $20.21\%$ & $29.27\%$ \\ 
\hline
$9$ & kNN & MLP w/ SeLU & $97.75\%$ & $97.56\%$ & $43.27\%$ & $94.66\%$ & $10.44\%$ & $11.62\%$ \\ 
\hline
$9$ & kNN & MLP w/ Leaky ReLU & $98.09\%$ & $97.56\%$ & $92.34\%$ & $95.71\%$ & $36.16\%$ & $35.49\%$ \\ 
\hline
$9$ & kNN & MLP w/ ReLU & $98.03\%$ & $97.62\%$ & $92.65\%$ & $95.53\%$ & $37.03\%$ & $34.39\%$ \\ 
\hline
\hline
\multicolumn{3}{|@{\hskip 1pt}c@{\hskip 2pt}|}{CNN}  & $96.47\%$ & $96.9\%$ & $95.06\%$ & $95.51\%$ & $75.71\%$ & $74.28\%$ \\ 
\hline
\multicolumn{3}{|@{\hskip 1pt}c@{\hskip 2pt}|}{Deformable CNN}  & $98.21\%$ & $98.60\%$ & $96.59\%$ & $97.60\%$ & $80.82\%$ & $81.01\%$ \\ 
\hline
\multicolumn{3}{|c|}{CapsNet} & $\mathbf{99.8}\%$ & $\mathbf{99.9}\%$ & $\mathbf{99.6}\%$ & $\mathbf{99.5}\%$ & $\mathbf{92.5}\%$ & $83.8\%$ \\ 
\hline
\end{tabular}
\end{center}
\caption{MNIST performance comparison with different rotations under the setting of farthest subsampling. P stands for polynomials, and NBR stands for neighbor. The first row indicates the rotation angles of test images.}
\label{table:MNISTRotateResults}
\end{table*}

\noindent \textbf{Robustness to Rotations}
The next experiment we conduct is to train with objects of limited known rotations and test on objects with various rotated angles. We define counter clockwise as the positive rotation direction. The training set is constructed through rotating the original $28\times 28$ training images (or point clouds) by $-15^\circ$, $0^\circ$, and $+15^\circ$ around the center. We rotate the original testing images (or point clouds) by $-15^\circ$ and $+15^\circ$ to create the validation set. All parameters are tuned on the validation dataset. Test accuracies are measured on the rotation angles of $\pm 10^\circ$, $\pm 20^\circ$, and $\pm 40^\circ$.

Trends on results shown in Table \ref{table:MNISTRotateResults} are similar to those mentioned in the previous subsection, e.g. $\epsilon$-ball and cubic polynomials outperform kNN and MLPs. However, here the regular ReLU seems to show the most robustness among the activation functions, although cubic polynomials still outperform all the MLP-based WeightNets. With $\epsilon$-ball and cubic polynomials, PointConv is again significantly more robust than regular CNNs. Interestingly CapsNet is able to show better rotation robustness than PointConv. Given that CapsNet is orthogonal to PointConv, it maybe possible to combine them in future work.

\begin{table*}[t!]
\begin{center}
\small
\begin{tabular}{ c | c } 
 \hline
 \textbf{Layer name} & \textbf{Layer description}  \\ 
 \hline
 Conv1 & $3\times3$ conv. (or PointConv) 64 w/ BN \& ReLU\\
 \hline
 Subsampling1 & max-pooling with stride $2\times 2$ (or farthest point sampling)\\
 \hline
 Conv2 & $3\times3$ conv. (or PointConv) 128 w/ BN \& ReLU\\
 \hline
 Conv3 & $3\times3$ conv. (or PointConv) 128 w/ BN \& ReLU\\
 \hline
 Subsampling2 & max-pooling with stride $2\times 2$ (or farthest point sampling)\\
 \hline
 Conv4 & $3\times3$ conv. (or PointConv) 256 w/ BN \& ReLU\\
 \hline
 Conv5 & $3\times3$ conv. (or PointConv) 256 w/ BN \& ReLU\\
 \hline
Subsampling3 & max-pooling with stride $2\times 2$ (or farthest point sampling)\\
 \hline
 Conv6 & $3\times3$ conv. (or PointConv) 512 w/ BN \& ReLU\\
 \hline
 Conv7 & $3\times3$ conv. (or PointConv) 512 w/ BN \& ReLU\\
\hline
Pooling & global average pooling\\
 \hline
FC-10 & fully connected layer\\
 \hline
\multicolumn{2}{c}{Softmax layer}\\
 \hline
\end{tabular}
\end{center}
\caption{The general network architecture for CIFAR-10 experiments. BN stands for batch normalization. Every BN layer is followed by a ReLU layer.}
\label{table:CIFAR_arch}
\end{table*}

\subsection{CIFAR-10}
In CIFAR-10~\cite{Cifar}, there are $60,000$ RGB images in $10$ classes in the training set, and $10,000$ images in the test set. The original size of each image is $32\times32$. The general architecture of all baselines are shown in Table \ref{table:CIFAR_arch}. We adopt the same hyperparameters and preprocessing procedures as detailed in section \ref{PConv_MNIST_Experiments}, besides the Sobolev norm regularizer. The $\epsilon$ is fixed to $0.05$ for Conv1 in Table \ref{table:CIFAR_arch}, $0.1$ for Conv2 and Conv3, $0.2$ for Conv4 and Conv5, and $0.5$ for Conv6 as well as Conv7. The numbers of output points from FPS for Subsampling1, Subsampling2, and Subsampling3 layer in Table \ref{table:CIFAR_arch} are $256$, $64$, and $16$, respectively. The coefficient of the Sobolev norm regularization is set to one of $\{10^{-6}, 10^{-5}, 10^{-4}\}$ for our proposed approach. We does not adopt PC-2D subsampling since it is experimentally proven not robust in \ref{PConv_MNIST_Experiments}. 

Two 2D CNN baselines are considered. The first one is the traditional CNN as shown in Table \ref{table:CIFAR_arch}, and the other one is Deformable CNN \cite{Dai2017} which has the exact architecture except for that the last four convolutional layer are deformable.We did not including CapsNet \cite{DBLP:journals/corr/abs-1710-09829} in this experiment, since it has no pooling layers and performs poorly on this dataset.

\noindent\textbf{Robustness to Scaling}
The first experiment tests robustness with respect to scaling. The original training set is augmented to three different resolutions, which are $24\times24$, $32\times32$, and $40\times40$, respectively. The validation resolutions are $28\times28$, and $36\times36$. We trained all baseline models on this new CIFAR-10 dataset, and tuned hyperparameters on the validation set.

Results are shown in Table~\ref{table:CIFARFarthestSubResults}. We note several takeaways:
\begin{itemize}
    \item Naive kNN-based PointConv is less robust to scale changes than conventional CNN. On most out-of-sample scales it does not generalize as well as the 2D CNNs. Even for the resolution $32\times 32$, the generalization is less well than traditional CNNs. This is interesting since we have verified that with the same architecture PointConv is able to match CNN performance if trained only under a single resolution, which might show that further work is needed for PointConv to be robust to mixing resolutions in training.
    \item Cubic polynomials outperform MLP with any activation at most scales. Sobolev regularization does not universally improves the robustness against scaling on this dataset. 
    \item $\epsilon$-ball neighborhood is generally significantly more robust than kNN neighborhood when testing scales are significantly different from the training.
    \item With appropriate design e.g. cubic polynomial WeightNet and an $\epsilon$-ball neighborhood, PointConv is significantly more robust than traditional 2D CNN when the testings scales that are significantly larger/smaller than training scales.
\end{itemize}

\begin{table*}[t!]
\begin{center}
\small
\begin{tabular}{|@{\hskip 1pt}c@{\hskip 1pt}|@{\hskip 1pt}c@{\hskip 1pt}|@{\hskip 2pt}c@{\hskip 2pt}|@{\hskip 2pt} c@{\hskip 2pt} |@{\hskip 2pt} c@{\hskip 2pt} |@{\hskip 2pt} c@{\hskip 2pt} |@{\hskip 2pt} c@{\hskip 2pt} |@{\hskip 2pt} c@{\hskip 2pt} ||@{\hskip 2pt} c @{\hskip 2pt}| @{\hskip 2pt}c @{\hskip 2pt}| @{\hskip 2pt}c @{\hskip 2pt}| } 
 \hline
 \textbf{\# NBR} & \textbf{Neighborhood} & \textbf{WeightNet architecture}  & $32\times32$ & $48\times48$ & $60\times60$ & $76\times76$ & $88\times88$ & $22\times22$ & $18\times18$ & $16\times16$  \\ 
\hline
$16$ & $\epsilon$-ball & cubic P + Sobolev Reg.& $73.43\%$ & $70.24\%$ & $\mathbf{68.56}\%$ & $\mathbf{65.7}\%$ & $\mathbf{65.78}\%$ & $66.15\%$ & $\mathbf{67.12}\%$ & $\mathbf{71.89}\%$  \\ 
\hline
$9$ & $\epsilon$-ball & cubic P + Sobolev Reg. & $70.56\%$ & $68.00\%$ & $67.15\%$ & $64.41\%$ & $65.31\%$ & $59.89\%$ & $63.36\%$ & $66.41\%$  \\ 
\hline
\hline
$9$ & kNN & cubic P + Sobolev Reg. & $78.59\%$ & $61.67\%$ & $46.32\%$ & $32.48\%$ & $24.91\%$ & $64.67\%$ & $52.50\%$ & $59.65\%$  \\ 
\hline
$9$ & kNN & cubic P  & $78.09\%$ & $61.87\%$ & $47.12\%$ & $30.91\%$ & $27.63\%$ & $63.17\%$ & $54.48\%$ & $59.65\%$  \\ 
\hline
$9$ & kNN & MLP w/ Sine   & $71.29\%$ & $39.62\%$ & $30.58\%$ & $26.69\%$ & $25.17\%$ & $46.68\%$ & $21.99\%$ & $23.81\%$  \\ 
\hline
$9$ & kNN & MLP w/ SeLU & $72.04\%$ & $47.27\%$ & $37.48\%$ & $34.81\%$ & $34.54\%$ & $35.21\%$ & $18.51\%$ & $23.80\%$  \\ 
\hline
$9$ & kNN & MLP w/ Leaky ReLU  & $76.06\%$ & $16.85\%$ & $9.85\%$ & $11.37\%$ & $11.21\%$ & $41.56\%$ & $30.11\%$ & $27.47\%$  \\ 
\hline
$9$ & kNN & MLP w/ ReLU  & $69.58\%$ & $20.81\%$ & $15.00\%$ & $12.01\%$ & $12.36\%$ & $38.89\%$ & $27.17\%$ & $25.32\%$  \\ 
\hline
\hline
\multicolumn{3}{|@{\hskip 1pt}c@{\hskip 2pt}|}{CNN} & $\mathbf{88.47}\%$ & $\mathbf{79.21}\%$ & $59.21\%$ & $38.43\%$ & $27.13\%$ & $\mathbf{69.11}\%$ & $44.99\%$ & $39.80\%$  \\ 
\hline
\multicolumn{3}{|@{\hskip 1pt}c@{\hskip 2pt}|}{Deformable CNN} & $82.82\%$ & $72.03\%$ & $52.07\%$ & $31.91\%$ & $25.68\%$ & $60.20\%$ & $51.20\%$ & $45.67\%$  \\ 
\hline
\end{tabular}
\end{center}
\caption{CIFAR-10 performance comparison across different scales under the setting of farthest subsampling. P is short for polynomials, and NBR is short for neighbor. The first row indicates the resolution of test images.}
\label{table:CIFARFarthestSubResults}
\end{table*}

\begin{table*}[h!]
\begin{center}
\small
\begin{tabular}{|@{\hskip 1pt} c @{\hskip 1pt}|@{\hskip 1pt} c @{\hskip 1pt}|@{\hskip 1pt} c @{\hskip 1pt}|@{\hskip 1pt} c @{\hskip 1pt}|@{\hskip 1pt} c @{\hskip 1pt}|@{\hskip 1pt} c @{\hskip 1pt}|@{\hskip 1pt} c @{\hskip 1pt}|@{\hskip 1pt} c @{\hskip 1pt}|@{\hskip 1pt} c @{\hskip 1pt}|@{\hskip 1pt} c @{\hskip 1pt}| } 
 \hline
 \textbf{\# NBR} & \textbf{Neighborhood} & \textbf{WeightNet architecture} & $0\degree$ & $+10\degree$ & $-10\degree$ & $+20\degree$ & $-20\degree$ & $+40\degree$ & $-40\degree$ \\ 
\hline
$12$ & $\epsilon$-ball & cubic P + Sobolev Reg. & $71.26\%$ & $67.63\%$ & $67.55\%$ & $64.58\%$ & $65.36\%$ & $40.13\%$ & $42.64\%$ \\ 
\hline
$9$ & $\epsilon$-ball & cubic P + Sobolev Reg. & $69.72\%$ & $66.24\%$ & $66.56\%$ & $63.13\%$ & $63.71\%$ & $39.42\%$ & $39.42\%$\\ 
\hline
\hline
$9$ & kNN & cubic P + Sobolev Reg. & $75.57\%$ & $62.95\%$ & $60.12\%$ & $59.96\%$ & $54.35\%$ & $42.94\%$ & $39.13\%$ \\ 
\hline
$9$ & kNN & cubic P & $75.94\%$ & $62.27\%$ & $59.12\%$ & $57.18\%$ & $54.68\%$ & $38.85\%$ & $37.61\%$ \\ 
\hline
$9$ & kNN & MLP w/ Sine & $68.32\%$ & $45.11\%$ & $44.11\%$ & $35.92\%$ & $35.52\%$ & $15.35\%$ & $16.34\%$ \\ 
\hline
$9$ & kNN & MLP w/ SeLU & $69.04\%$ & $43.81\%$ & $41.26\%$ & $40.83\%$ & $37.97\%$ & $21.53\%$ & $18.28\%$ \\ 
\hline
$9$ & kNN & MLP w/ Leaky ReLU & $78.01\%$ & $65.68\%$ & $64.84\%$ & $59.22\%$ & $57.85\%$ & $34.67\%$ & $34.14\%$ \\ 
\hline
$9$ & kNN & MLP w/ ReLU  & $76.79\%$ & $64.57\%$ & $62.25\%$ & $57.43\%$ & $55.21\%$ & $31.07\%$ & $33.98\%$ \\ 
\hline
\hline
\multicolumn{3}{|@{\hskip 1pt}c@{\hskip 2pt}|}{CNN}  & $\mathbf{86.61}\%$ & $83.72\%$ & $\mathbf{84.48}\%$ & $\mathbf{78.03}\%$ & $78.25\%$ & $45.71\%$ & $46.14\%$ \\ 
\hline
\multicolumn{3}{|@{\hskip 1pt}c@{\hskip 2pt}|}{Deformable CNN} & $86.53\%$ & $\mathbf{83.74}\%$ & $83.96\%$ & $77.69\%$ & $\mathbf{78.31}\%$ & $\mathbf{47.21}\%$ & $\mathbf{47.72}\%$ \\ 
\hline
\end{tabular}
\end{center}
\caption{CIFAR-10 performance comparison with different rotations under the setting of farthest subsampling. P stands for polynomials, and NBR stands for neighbor. The first row indicates the rotation angles of test images.}
\label{table:CIFARRotateResults}
\end{table*}

\noindent \textbf{Robustness to Rotations}
The next experiment we conduct is to train baselines with images with a few predefined rotations, and test them on images with various rotated angles. The counter clockwise direction is defined as the positive direction. To construct the training set, we rotate the original $32\times32$ training images (or point clouds) by $-15^\circ$, $0^\circ$, and $+15^\circ$ around the center. The validation set is created through rotating the original testing images (or point clouds) by $-15^\circ$ and $+15^\circ$. All parameters are tuned on the validation dataset. Test accuracies are measured on the rotation angles of $\pm 10^\circ$, $\pm 20^\circ$, and $\pm 40^\circ$.

The corresponding results are shown in Table \ref{table:CIFARRotateResults}. We observe that $\epsilon$-ball still outperforms kNN in all nonzero rotations but underperform it at no rotation. However, Leaky ReLU is overall comparable with cubic polynomial with Sobolev regularization as the best approach, and SeLU performed significantly worse. Besides, PointConv is generally significantly less robust as regular or deformable CNNs in this case, showing that more work needs to be done in terms of robustness to rotations in 2D in more complex datasets. 

\begin{table*}[h!]
\begin{center}
\small
\begin{tabular}{| c | c | c | c | c | c | c | c | c | c | } 
\hline
\textbf{\# layers} & \textbf{MLP input} & \textbf{activation} & \textbf{NBR setting} & \makecell{\textbf{rotation} \\ \textbf{augmentation}} & $100k$ & $60k$ & $40k$ & $20k$ & $10k$ \\ \hline
16 & VI + $(x,y,z)$ & ReLU & KNN & Y & $\mathbf{68.2}$ &  $\mathbf{64.4}$ & $\mathbf{63.0}$ & $\mathbf{57.6}$ &  $45.4$\\
\hline
16 & VI & ReLU & KNN & Y & $63.3$ &  $60.7$ & $59.7$ & ${55.0}$ &  $44.7$\\
\hline
16 & VI & SeLU & KNN & Y & ${63.7}$ & ${61.5}$  & $57.7$ & $53.1$ & $40.2$ \\
\hline
\hline
16 & VI & Cubic First Block & KNN & Y & $63.5$ & $61.4$  & ${60.2}$ & $54.0$ & $42.0$\\
& &+ ReLU in others  & & & & & & & \\
\hline
16 & $(x,y,z)$ & ReLU & KNN & Y & $61.7$ &  $58.7$ & $53.4$  & $34.6$ & $17.8$\\
\hline
16 & VI & ReLU & KNN & N & $60.4$ &  $58.6$ & $57.8$ & $54.2$ &  $46.3$\\
\hline
16 & $(x,y,z)$ & ReLU & KNN & N & $56.4$ & $54.0$ & $51.2$ & $40.6$ & $27.4$ \\
\hline
\hline
16 & VI & ReLU & $\epsilon$-ball & Y & $61.61$ &  $58.6$ & $58.0$ & $52.3$ & $40.6$ \\
\hline
16 & $(x,y,z)$ & ReLU & $\epsilon$-ball & Y  & $48.9$ &  $43.3$ & $39.7$ & $30.6$ & $20.7$ \\
\hline
\hline
16 & surface normal & ReLU & KNN & Y & $60.2$ &  $57.5$ & $56.8$ & $54.8$ &  $50.9$ \\
& +  $(x,y,z)$ & & & & & & & &  \\
\hline
16 & surface normal & ReLU & KNN & Y & $53.1$ &  $50.6$ & $50.2$ & $47.6$ &  $43.3$\\
\hline
\hline
4 & VI + $(x,y,z)$ & ReLU & KNN & Y & $64.5$ &  $61.3$ & $60.6$  & $57.3$ & $\mathbf{51.2}$\\
\hline
4 & VI  & ReLU & KNN & Y  & $61.0$ &  $58.8$ & $57.5$  & $50.8$ & $39.4$\\
\hline
4 & $(x,y,z)$ & ReLU & KNN & Y & $55.3$ &  $53.3$ & $47.0$  & $30.7$ & $16.1$\\
\hline
4 & VI  & ReLU &  $\epsilon$-ball & Y & $59.2$ &  $57.5$ & $55.12$  & $44.8$ & $31.1$\\
\hline
\end{tabular}
\end{center}
\vskip -0.05in
\caption{Performance results (mIoU,\%) for the ScanNet dataset. The first column shows the configurations of the approach, and the top row contains the number of subsampled points. The default number of neighbors is $8$, and the default activation for weight functions is ReLU.}
\label{table:VIScanNet}
\vskip -0.05in
\end{table*}

\subsection{ScanNet}
We conduct 3D semantic scene segmentation on the ScanNet v2 \cite{dai2017scannet} dataset. 
We use the official split with $1,201$ scenes for training and $312$ for validation.

We implemented 2 PointConv architectures. One is the 4-layer network in~\cite{Wu_2019_CVPR}, the second one is the 16-layer PointConv network that achieved $66.6\%$ on the ScanNet testing set. Network architectures are provided by the authors. Our main results are reported on the ScanNet validation set as the benchmark organizers do not allow ablation studies on the testing set. 

We adopt regular subsampling \cite{8490990} for feature encoding layers with grid sizes $0.05$, $0.1$, $0.2$, and $0.4$ (in meter). $\epsilon$ is set to be $\frac{1}{1.3}$ of the grid size for the corresponding subsampling layer. We enumerated $\epsilon$ in $\{\frac{1}{1.0},\frac{1}{1.1},\frac{1}{1.2},\frac{1}{1.3},\frac{1}{1.4},\frac{1}{1.5}\}$, and it turned out the network is not sensitive to those choices (see supplementary for experiments). The surface normal for a subsampled point is computed through averaging all surface normals of its corresponding grid. During training, we randomly subsample $100,000$ points from each point cloud for both the training and validation sets, and the mini-batch size is set to be $3$. The learning rate is set to $10^{-3}$, with a decay multiplier of $\frac{1}{2}$ every $40$ epochs. Moreover, the rotation augmentation is applied by randomly rotating every mini-batch with an arbitrary angle in $[0, 2\pi)$, as in~\cite{Wu_2019_CVPR}.

To study the robustness to scales for all baselines, we re-subsample each validation point cloud to less than $100k$ --- $\{60k,40k,20,10k\}$. This is equivalent to downsampling the image in the 2D space as it increases the size of KNN neighborhoods with a fixed $K$. Also, each sub-sampled point cloud is further rotated with $4$ different predefined angles around $z$-axis --- $\{0\degree, 90\degree, 180\degree, 270\degree\}$. Such operations could significantly change both local scales and rotations. We also evaluate the performance when the rotation augmentation is not applied during training. 
We found performance variation between different rotation angles to be less than $1\%$ (see supplementary), hence the mIoUs averaged from all angles are reported.
Experiments are performed with the proposed VI descriptor, as well as configurations that are promising from 2D experiments: $\epsilon$-ball, cubic polynomial WeightNet, and SeLU activation.

Results are shown in Table \ref{table:VIScanNet}, which shows that the proposed VI descriptor significantly improved the performance as well as robustness under every setting. Especially, it is significantly more robust to input downsampling than the $(x,y,z)$ coordinates as input. For example, at $10k$ testing points (reflecting $10x$ downsampling from the training), the VI descriptors still maintain a $44.7\%$ accuracy while the $(x,y,z)$ coordinates version has its performance dropped to $17.8\%$, marking an improvement of $251\%$. Besides, with VI descriptor, the need to use rotation augmentation reduced significantly.  Interestingly, rotation augmentation still improved performance by $2\%$. We believe the main reason is that the input feature of our framework consists of both $(x,y,z)$ coordinates and RGB colors, and thus rotation augmentation creates more  $(x,y,z)$ coordinates variations for the training, and hence improved  performance. We further conduct ablation studies by replacing VI descriptors with surface normals. The mIoU drops by $8\%-10\%$, which indicates that VI is a better representation of local geometry than the commonly used surface normal.

Finally, when we combined the VI features with $(x,y,z)$ inputs, it generated the best performance of all -- $68.2\%$ on the original validation set, and better on almost all subsampled scenarios. This shows that a combination of scale-invariant, rotation-invariant and non-invariant features is beneficial, potentially letting the network choose the invariance it requires. Furthermore, on the test set, we achieved comparable mIoU with KPConv \cite{thomas2019KPConv}, which is the current state-of-the-art among point-based approaches (Table \ref{table:ScannetTest}). Note that PointConv \cite{Wu_2019_CVPR} has significantly less parameters than KPConv \cite{thomas2019KPConv}, and KNN used in PointConv is significantly more efficient than the $\epsilon$-ball in KPConv.

Otherwise, none of the other tested choices were helpful, including $\epsilon$-ball, SeLU or cubic polynomial. Our takeaway is that the proposed VI descriptor is very powerful and improved robustness significantly w.r.t.  neighborhood sizes, hence, none of the other improvements is needed.

From our experience, it is very difficult to find a good set of parameters for $\epsilon$-ball in $3D$. It also slows down the algorithm because the non-uniform neighborhood size is not easily amenable for tensor computation. Hence not needing to use it is a significant bonus. Nevertheless, $\epsilon$-ball neighborhoods are still very useful in $2D$ settings, where locating them is much easier and the VI descriptor is not applicable. 

\begin{table}[h!]
\begin{center}
\small
\begin{tabular}{ c | c } 
 \hline
 \textbf{Method}  &  \textbf{mIoU(\%)} \\ 
\hline
 PointNet++ \cite{qi2017pointnetplusplus}  &  $33.9$ \\ 
\hline
 SPLATNet \cite{su18splatnet} &  $39.3$  \\
\hline
 TangentConv \cite{Tat2018}  &  $40.9$ \\ 
\hline
 PointCNN \cite{NIPS2018_7362}  &  $45.8$ \\ 
\hline
 PointASNL \cite{Yan_2020_CVPR}  &  $63.0$ \\ 
\hline
 PointConv \cite{Wu_2019_CVPR}  &  $66.6$ \\ 
\hline
 KPConv \cite{thomas2019KPConv}  &  $\mathbf{68.4}$ \\ 
\hline
\hline
 VI-PointConv (ours) & $67.6$ \\
\hline
\end{tabular}
\end{center}
\vskip -0.05in
\caption{Semantic Scene Segmentation results for point-based approaches on the ScanNet test set}
\label{table:ScannetTest}
\vskip -0.05in
\end{table}

\begin{table*}[h!]
\begin{center}
\small
\begin{tabular}{ @{}c@{} | @{\hskip 2pt}c@{\hskip 2pt}| @{\hskip 2pt}c@{\hskip 2pt}|@{\hskip 2pt}c@{\hskip 2pt}|@{\hskip 2pt}c@{\hskip 2pt}|@{\hskip 2pt}c@{\hskip 2pt}|@{\hskip 2pt}c@{\hskip 2pt}|@{\hskip 2pt}c@{\hskip 2pt}|@{\hskip 2pt}c@{\hskip 2pt}|@{\hskip 2pt}c@{\hskip 2pt}|@{\hskip 2pt}c@{\hskip 2pt}|@{\hskip 2pt}c@{\hskip 2pt}|@{\hskip 2pt}c@{\hskip 2pt}|@{\hskip 2pt}c@{\hskip 2pt}|@{\hskip 2pt}c@{\hskip 2pt}|@{\hskip 2pt}c@{\hskip 2pt}|@{\hskip 2pt}c@{\hskip 2pt}|@{\hskip 2pt}c@{\hskip 2pt}|@{\hskip 2pt}c@{\hskip 2pt}|@{\hskip 2pt}c@{\hskip 2pt}|@{\hskip 2pt}c@{\hskip 2pt}} 
 \hline
 \textbf{Method} & \rotatebox{90}{\textbf{mIoUs(\%)}} & \rotatebox{90}{road}  &  \rotatebox{90}{sidewalk} & \rotatebox{90}{parking} & \rotatebox{90}{other-ground}  &  \rotatebox{90}{building} & \rotatebox{90}{car} & \rotatebox{90}{truck} &  \rotatebox{90}{bicycle} & \rotatebox{90}{motorcycle} & \rotatebox{90}{other-vehicle} & \rotatebox{90}{vegetation} & \rotatebox{90}{trunk} & \rotatebox{90}{terrain} & \rotatebox{90}{person} & \rotatebox{90}{bicyclist} & \rotatebox{90}{motorcyclist} & \rotatebox{90}{fence} &  \rotatebox{90}{pole} & \rotatebox{90}{traffic-sign}  \\ 
 
 \hline
PointNet\cite{qi2016pointnet} & $14.6$ & $61.6$ & $35.7$ & $15.8$ & $1.4$ & $41.4$ & $46.3$ & $0.1$ & $1.3$ & $0.3$ & $0.8$ & $31.0$ & $4.6$ & $17.6$ & $0.2$ & $0.2$ & $0.0$ & $12.9$ & $2.4$ & $3.7$  \\
 \hline
SPG\cite{8578577} & $17.4$ & $45.0$ & $28.5$ & $0.6$ & $0.6$ & $64.3$ & $49.3$ & $0.1$ & $0.2$ & $0.2$ & $0.8$ & $48.9$ & $27.2$ & $24.6$ & $0.3$ & $2.7$ & $0.1$ & $20.8$ & $15.9$ & $0.8$ \\
 \hline
SPLATNet\cite{su18splatnet} & $18.4$ & $64.6$ & $39.1$ & $0.4$ & $0.0$ & $58.3$ & $58.2$ & $0.0$ & $0.0$ & $0.0$ & $0.0$ & $71.1$ & $9.9$ & $19.3$ & $0.0$ & $0.0$ & $0.0$ & $23.1$ & $5.6$ & $0.0$ \\
 \hline
PointNet++\cite{qi2017pointnetplusplus}  & $20.1$ & $72.0$ & $41.8$ & $18.7$ & $5.6$ & $62.3$ & $53.7$ & $0.9$ & $1.9$ & $0.2$ & $0.2$ & $46.5$ & $13.8$ & $30.0$ & $0.9$ & $1.0$ & $0.0$ & $16.9$ & $6.0$ & $8.9$ \\
 \hline
TangentConv\cite{Tat2018} & $40.9$ & $83.9$ & $63.9$ & $33.4$ & $15.4$ & $83.4$ & $90.8$ & $15.2$ & $2.7$ & $16.5$ & $12.1$ & $79.5$ & $49.3$ & $58.1$ & $23.0$ & $28.4$ & $8.1$ & $49.0$ & $35.8$ & $28.5$ \\
 \hline
PointConv\cite{Wu_2019_CVPR} & $53.0$ & $86.2$ & $68.6$ & $57.7$ & $16.0$ & $89.9$ & $94.2$ & $30.2$ & $29.5$ & $33.9$ & $30.5$ & $78.9$ & $60.8$ & $63.7$ & $48.8$ & $45.7$ & $20.4$ & $59.9$ & $53.4$ & $38.6$   \\ 
 \hline
RandLA-Net\cite{hu2019randla} & $53.9$ & $\mathbf{90.7}$ & $\mathbf{73.7}$ & $60.3$ & $20.4$ & $86.9$ & $94.2$ & $40.1$ & $26.0$ & $25.8$ & $38.9$ & $81.4$ & $61.3$ & $66.8$ & $49.2$ & $48.2$ & $7.2$ & $56.3$ & $49.2$ & $47.7$ \\
 \hline
KPConv\cite{thomas2019KPConv}  & $58.8$ & $88.8$ & $72.7$ & $61.3$ & $31.6$ & $90.5$ & $\mathbf{96.0}$ & $33.4$ & $30.2$ & $\mathbf{42.5}$ & $44.3$ & $\mathbf{84.8}$ & $\mathbf{69.2}$ & $\mathbf{69.1}$ & $\mathbf{61.5}$ & $\mathbf{61.6}$ & $11.8$ & $64.2$ & $56.5$ & $47.4$  \\ 
 
\hline
\hline
 VI-PConv(ours) & $\mathbf{59.6}$ & $88.8$ & $72.5$ & $\mathbf{63.5}$ & $\mathbf{32.7}$ & $\mathbf{91.4}$ & $95.9$ & $\mathbf{41.8}$ & $\mathbf{38.6}$ & $35.0$ & $\mathbf{45.7}$ & $83.9$ & $68.0$ & $66.9$ & $51.2$ & $50.1$ & $\mathbf{27.6}$ & $\mathbf{66.6}$ & $\mathbf{57.4}$ & $\mathbf{54.8}$               \\
\hline
\end{tabular}
\end{center}
\vskip -0.05in
\caption{Semantic Scene Segmentation results for point-based approaches on the SemanticKITTI test set}
\label{table:semanticKITTI}
\vskip -0.1in
\end{table*}

\subsection{SemanticKITTI}
We also evaluate the semantic segmentation performance on SemanticKITTI \cite{behley2019iccv} (single scan), which consists of $43,552$ point clouds sampled from $22$ sequences in driving scenes. Each point cloud contains $10-13k$ points, collected by a single Velodyne HDL-64E laser scanner, spanning up to $160\times160\times20$ meters in 3D space. The officially training set includes $19,130$ scans (sequences $00-07$ and $09-10$), and there are $4,071$ scans (sequence $08$) for validation. For each 3D point, only $(x,y,z)$ coordinate is given without any color information. It is a challenging dataset because faraway points are sparser in LIDAR scans. 

We adopt the exact same $16$-layer architecture as showed at the first row of Table \ref{table:VIScanNet}, together with the mini-batch size of 16. The initial learning rate is $10^{-3}$, and it is decayed by half every $6$ epochs. We do not integrate with any subsampling preprocessing.  As reported in Table \ref{table:semanticKITTI}, we achieve the state-of-the-art semantic segmentation performance among point-based baselines, improving by $0.8\%$ over KPConv and $6.6\%$ over standard PointConv, which demonstrate the effectiveness of VI descriptor for PointConv \cite{Wu_2019_CVPR}.

\section{Conclusion}
This paper empirically studies several strategies to improve the robustness for point cloud convolution with PointConv \cite{Wu_2019_CVPR}. We have found two combinations to be effective. In 2D images, using an $\epsilon$-ball neighborhood with cubic polynomial weight functions achieved the highest robustness, and outperformed 2D CNNs for MNIST dataset. In 3D images, our novel viewpoint-invariant descriptor, when used instead of, or in combination with the 3D coordinates, significantly improved the performance and robustness of PointConv networks and achieved state-of-the-art. Our results show that kNN is still a viable neighborhood choice if the input location features are robust to neighborhood sizes.

{\small
\bibliographystyle{ieee_fullname}
\bibliography{egbib}
}

\end{document}


\maketitle
\subsection{MNIST Network Architecture}
\vskip -0.1in
Table \ref{table:MNIST arch} illustrate the general architecture for our proposed framework and all other baselines. They only differ in the structures for convolutional layers. 
We utilize cross entropy as the loss function. The batch size is $60$ and the optimizer is Adam with learning rate $0.001$.

\begin{table}[h!]
\begin{center}
\small
\begin{tabular}{ c | c } 
 \hline
 \textbf{Layer name} & \textbf{Layer description}  \\ 
 \hline
 Conv1 & $3\times3$ conv. (or PointConv) 64 w/ ReLU\\
 \hline
 Subsampling1 & max-pooling with stride $2\times 2$ (or farthest point sampling)\\
 \hline
 Conv2 & $3\times3$ conv. (or PointConv) 128 w/ ReLU\\
 \hline
 Subsampling2 & max-pooling with stride $2\times 2$ (or farthest point sampling)\\
 \hline
 Conv3 & $3\times3$ conv. (or PointConv) 128 w/ ReLU\\
 \hline
 Conv4 & $3\times3$ conv. (or PointConv) 128 w/ ReLU\\
 \hline
Pooling & global average pooling\\
 \hline
FC-10 & fully connected layer\\
 \hline
\multicolumn{2}{c}{Softmax layer}\\
 \hline
\end{tabular}
\end{center}
\vskip -0.15in
\caption{The general network architecture for MNIST experiments. Every convolutional layer is followed by a ReLU layer.}
\label{table:MNIST arch}
\vskip -0.3in
\end{table}

\subsection{$\epsilon$-ball and FPS parameters}
\vskip -0.1in
The $\epsilon$ is fixed to $\frac{1}{10}$ for Conv1 in Table \ref{table:MNIST arch}, $\frac{1}{5}$ for Conv2, and $\frac{1}{2}$ for Conv3 as well as Conv4. The numbers of output points from FPS for Subsampling1 Subsampling2 layer in Table \ref{table:MNIST arch} are $196$ and $49$ respectively. 

\subsection{Grid Subsampling in MNIST}
\vskip -0.1in
We present results using PC-2D subsampling, which would just subsample one point per each $2\times 2$ window. This is the same sampling approach as used in 2D raster CNNs. Table \ref{table:MNISTGridSubResults} shows that it is generally inferior to the furthest point subsampling we utilized in the main paper.

\begin{table}[t!]
\begin{center}
\scriptsize
\begin{tabular}{|@{\hskip 1pt}c@{\hskip 1pt}|@{\hskip 1pt}c@{\hskip 1pt}|@{\hskip 2pt}c@{\hskip 2pt}|@{\hskip 2pt} c@{\hskip 2pt} |@{\hskip 2pt} c@{\hskip 2pt} |@{\hskip 2pt} c@{\hskip 2pt} |@{\hskip 2pt} c@{\hskip 2pt} |@{\hskip 2pt} c@{\hskip 2pt} ||@{\hskip 2pt} c @{\hskip 2pt}| @{\hskip 2pt}c @{\hskip 2pt}| @{\hskip 2pt}c @{\hskip 2pt}| }
 \hline
 \textbf{\# NBR} & \textbf{Neighborhood} & \textbf{WeightNet architecture}  & $34\times34$ & $38\times38$ & $44\times44$ & $56\times56$ & $72\times72$ & $18\times18$ & $14\times14$ & $10\times10$  \\ 
\hline
$16$ & $\epsilon$-ball & cubic P + Sobolev Reg. & $95.83\%$ & $96.21\%$ & $\mathbf{97.84}\%$ & $\mathbf{97.92}\%$ & $\mathbf{98.23}\%$ & $90.8\%$ & $64.47\%$ & $21.78\%$  \\ 
\hline
$9$ & $\epsilon$-ball & cubic P + Sobolev Reg. & $94.76\%$ & $95.58\%$ & $96.56\%$ & $96.89\%$ & $97.27\%$ & $\mathbf{93.51}\%$ & $\mathbf{88.20}\%$ & $\mathbf{23.59}\%$  \\ 
\hline
\hline
$9$ & KNN & cubic P + Sobolev Reg. & $97.64\%$ & $96.83\%$ & $95.92\%$ & $54.06\%$ & $23.99\%$ & $53.84\%$ & $17.98\%$ & $10.11\%$  \\ 
\hline
$9$ & KNN & cubic P & $97.85\%$ & $97.31\%$ & $95.38\%$ & $42.71\%$ & $13.77\%$ & $14.77\%$ & $12.02\%$ & $7.28\%$  \\ 
\hline
$9$ & KNN & MLP w/ Sine  & $95.76\%$ & $94.35\%$ & $94.47\%$ & $47.26\%$ & $19.02\%$ & $76.5\%$ & $30.01\%$ & $10.86\%$  \\ 
\hline
$9$ & KNN & MLP w/ SeLU & $92.03\%$ & $82.49\%$ & $56.90\%$ & $17.09\%$ & $6.01\%$ & $33.78\%$ & $10.51\%$ & $11.59\%$  \\ 
\hline
$9$ & KNN & MLP w/ Leaky ReLU & $66.27\%$ & $24.39\%$ & $13.69\%$ & $10.11\%$ & $10.1\%$ & $65.66\%$ & $16.50\%$ & $8.11\%$  \\ 
\hline
$9$ & KNN & MLP w/ ReLU & $16.54\%$ & $16.66\%$ & $11.35\%$ & $11.35\%$ & $11.35\%$ & $10.63\%$ & $7.83\%$ & $6.43\%$  \\ 
\hline
\hline
\multicolumn{3}{|@{\hskip 1pt}c@{\hskip 2pt}|}{CNN} & $96.30\%$ & $95.68\%$ & $89.19\%$ & $51.23\%$ & $25.77\%$ & $75.6\%$ & $24.25\%$ & $11.2\%$  \\ 
\hline
\multicolumn{3}{|@{\hskip 1pt}c@{\hskip 2pt}|}{Deformable CNN} & $91.31\%$ & $16.05\%$ & $17.99\%$ & $11.56\%$ & $11.35\%$ & $74.40\%$ & $28.18\%$ & $14.74\%$  \\ 
\hline
\multicolumn{3}{|c|}{CapsNet}   & $\mathbf{98.53}\%$ & $\mathbf{97.52}\%$ & $93.28\%$ & $43.25\%$ & $16.50\%$ & $80.07\%$ & $21.41\%$ & $8.65\%$  \\ 
\hline
\end{tabular}
\end{center}
\vskip -0.15in
\caption{MNIST  performance  comparison for detecting large/small objects, where P is short for polynomials, and NBR is short for neighbor. All methods in this table adopt PC-2D subsampling. The first row indicates the resolution of test images. Definitions of validation and test set are found in the paper.}
\label{table:MNISTGridSubResults}
\vskip -0.15in
\end{table}

\begin{table}[h!]
\captionsetup{font=scriptsize, width=.45\linewidth}
\begin{minipage}{.50\linewidth}
\centering
\scriptsize
\begin{tabular}{|c|c|c|c|c|c|}
\hline
Rotation angle  & $60k$ & $40k$ & $20k$ & $10k$ & $5k$ \\
\hline
0\degree & $60.8$ & $59.7$ & $55.3$ & $44.4$  & $32.5$ \\
\hline
90\degree  & $60.8$ & $59.7$ & $54.8$ & $44.7$ & $32.1$ \\
\hline
180\degree  & $60.3$ & $59.6$ & $54.7$ & $44.7$ & $32.4$\\
\hline
270\degree  & $60.8$ & $59.6$ & $55.2$ & $45.0$ & $32.0$ \\
\hline
\end{tabular}
\vskip -0.1in
\caption{Performance results (mIoU,\%) for the settings of KNN, VI descriptors as inputs for the MLP, and the network is trained with rotation augmentation.}
\label{table:KNN8DAug}
\end{minipage}%
\begin{minipage}{.50\linewidth}
\centering
\scriptsize
\begin{tabular}{|c|c|c|c|c|c|}
\hline
Rotation angle  & $60k$ & $40k$ & $20k$ & $10k$ & $5k$ \\
\hline
0\degree & $58.7$ & $58.0$ & $54.5$ & $46.1$ & $36.3$\\
\hline
90\degree & $58.7$ & $58.0$ & $53.6$ & $46.0$ & $36.3$\\
\hline
180\degree & $59.0$ & $57.2$ & $54.1$ & $46.6$ & $35.6$\\
\hline
270\degree & $58.1$ & $57.9$ & $54.5$ & $46.3$ &  $36.2$\\
\hline
\end{tabular}
\vskip -0.1in
\caption{Performance results (mIoU,\%) for the same settings as Table \ref{table:KNN8DAug}, expect that the network is trained without rotation augmentation.}
\label{table:KNN8DNoAug}
\end{minipage} 
\vskip -0.1in
\end{table}

\begin{table}[h!]
\captionsetup{font=scriptsize, width=.45\linewidth}
\begin{minipage}{.50\linewidth}
\centering
\scriptsize
\begin{tabular}{|c|c|c|c|c|c|}
\hline
Rotation angle  & $60k$ & $40k$ & $20k$ & $10k$ & $5k$\\
\hline
0\degree  & $58.8$ & $53.5$ & $35.0$ & $17.9$ & $10.0$ \\
\hline
90\degree & $58.6$ & $53.2$ & $34.3$ & $17.6$ & $9.9$\\
\hline
180\degree  & $58.6$ & $53.4$ & $34.6$ & $17.9$ & $10.0$\\
\hline
270\degree  & $58.8$ & $53.5$ & $34.5$ & $17.8$ & $9.9$\\
\hline
\end{tabular}
\vskip -0.1in
\caption{Performance results (mIoU,\%) for the settings of KNN, $(x,y,z)$ coordinates as inputs for the MLP, and the network is trained with rotation augmentation.}
\label{table:KNNXYZAug}
\end{minipage}%
\begin{minipage}{.50\linewidth}
\centering
\scriptsize
\begin{tabular}{|c|c|c|c|c|c|}
\hline
Rotation angle  & $60k$ & $40k$ & $20k$ & $10k$ & $5k$\\
\hline
0\degree & $54.2$ & $51.1$ & $40.4$ & $27.3$  & $16.1$\\
\hline
90\degree & $53.8$ & $50.9$ & $40.6$ & $27.4$ & $15.7$\\
\hline
180\degree & $54.0$ & $51.1$ & $40.6$ & $27.3$ & $16.1$\\
\hline
270\degree & $54.1$ & $51.6$ & $40.6$ & $27.6$ & $16.1$\\
\hline
\end{tabular}
\vskip -0.1in
\caption{Performance results (mIoU,\%) for the same settings as Table \ref{table:KNNXYZAug}, expect that the network is trained without rotation augmentation.}
\label{table:KNNXYZNoAug}
\end{minipage}
\vskip -0.1in
\end{table}

\begin{table}[t!]
\captionsetup{font=scriptsize, width=.45\linewidth}
\begin{minipage}{.50\linewidth}
\centering
\scriptsize
\begin{tabular}{|c|c|c|c|c|c|}
\hline
Rotation angle  & $60k$ & $40k$ & $20k$ & $10k$ & $5k$\\
\hline
0\degree  & $64.1$ & $63.0$ & $57.3$ & $45.4$ & $40.2$ \\
\hline
90\degree & $64.4$ & $62.8$ & $57.4$ & $45.3$ & $40.4$\\
\hline
180\degree  & $64.4$ & $63.5$ & $57.9$ & $45.6$ & $40.1$\\
\hline
270\degree  & $64.6$ & $62.7$ & $57.6$ & $45.3$ & $40.1$\\
\hline
\end{tabular}
\vskip -0.1in
\caption{Performance results (mIoU,\%) for the settings of KNN, VI + $(x,y,z)$ coordinates as inputs for the MLP, and the network is trained with rotation augmentation.}
\label{table:KNNXYZAug}
\end{minipage}%

\end{table}

\subsection{Extra Performance Reports for the ScanNet experiment}
\vskip -0.1in
We provide more detailed results for the ScanNet experiment, starting from Table \ref{table:KNN8DAug}. The first row of each Table indicates the number of subsampled points, and the first column represents rotation angles. Note that the performance differences under each column are not significant, which indicates that frameworks are robust against rotations. However, the performance decreases as the number of sampled points gets smaller and smaller. When the 3D coordinates input is replaced with our novel viewpoint-invariant descriptor, the performance and robustness are significantly improved for PointConv networks. To investigate the sensitivity of $\epsilon$, we tried $\{\frac{1}{1.0},\frac{1}{1.1},\frac{1}{1.2},\frac{1}{1.3},\frac{1}{1.4},\frac{1}{1.5}\}$, and the corresponding mIOUs are $\{62.0\%, 62.0\%, 61.9\%, 61.6\%, 61.6\%, 61.6\%\}$. To study the effectiveness of the VI descriptor on other frameworks, we replace edge inputs with VI descriptors on DGCNN\cite{dgcnn}, and the mIOU on ScanNet validation set improved by about $4\%$. It justified that the VI descriptor is quite useful in other point cloud networks as well. Other prior work, e.g. PointNet, didn’t consider point-point relations hence can’t use it. Since a proper CNN (e.g. PointConv) outperforms most of these non-convolutional networks on point clouds, we felt it’s more important to study the CNN.

{\small
\bibliographystyle{ieee_fullname}
\bibliography{egbib}
}